\documentclass[conference]{IEEEtran}

\usepackage{amsmath,amssymb,amsfonts}
\usepackage{algorithm}
\usepackage{algpseudocode}
\usepackage{booktabs}
\usepackage{multirow}
\usepackage{graphicx}
\usepackage{textcomp}
\usepackage{xcolor}
\usepackage{balance}
\usepackage{fancyhdr}
\usepackage{url}
\usepackage{hyperref}
\usepackage[square,comma,sort&compress,numbers]{natbib}
\usepackage{threeparttable,multirow,adjustbox,threeparttable}
\usepackage{tabularx}
\newcolumntype{Y}{>{\centering\arraybackslash}X}

\begin{document}
\title{
Style-Aware Blending and Prototype-Based Cross-Contrast Consistency for Semi-Supervised Medical Image Segmentation
}
\author{
    \IEEEauthorblockN{
        Chaowei Chen$^1$, Xiang Zhang$^1$, Honglie Guo$^1$, and Shunfang Wang$^{1,2,*}$
    } 
    \\
    \IEEEauthorblockA{
        $^1$School of Information Science and Engineering, Yunnan University, Kunming  650504, Yunnan, China \\
        $^2$Yunnan Key Laboratory of Intelligent Systems and Computing, Yunnan University, Kunming, 650504, Yunnan, China \\
        $^{*}$Corresponding author: \href{mailto:sfwang_66@ynu.edu.cn}{sfwang\_66@ynu.edu.cn}
    }
}

\maketitle
\thispagestyle{fancy}

\begin{abstract}
Weak-strong consistency learning strategies are widely employed in semi-supervised medical image segmentation to train models by leveraging limited labeled data and enforcing weak-to-strong consistency. However, most existing methods primarily focus on designing and combining various perturbation schemes, overlooking the intrinsic potential and limitations of the framework itself. In this paper, we identify two critical deficiencies: (1) separated training data streams, which lead to confirmation bias dominated by the labeled stream; and (2) incomplete utilization of supervisory signals, which limits exploration of strong-to-weak consistency. To address these challenges, we propose a style-aware blending and prototype-based cross-contrast consistency learning framework. Specifically, inspired by the empirical observation that the distribution mismatch between labeled and unlabeled data can be characterized by their statistical moments, we design a style-guided distribution blending module to bridge the independent training data streams. Meanwhile, considering the potential noise in strong pseudo-labels, we introduce a prototype-based cross-contrast strategy to enable the model to learn informative supervisory signals from both weak-to-strong and strong-to-weak predictions, while mitigating the adverse effects of noise. Extensive experiments demonstrate the effectiveness and superiority of our framework across multiple medical image segmentation benchmarks under various semi-supervised settings. The code is available at \href{https://gndlwch2w.github.io/spc-demo}{\textcolor[HTML]{E74697}{https://gndlwch2w.github.io/spc-demo}}. 
\end{abstract}

\begin{IEEEkeywords}
Semi-supervised learning, Style-guided distribution blending, Prototype-based cross-contrast learning
\end{IEEEkeywords}

\section{Introduction}
Accurate and cost-effective medical image segmentation is crucial for computer-aided clinical diagnosis \cite{liu2017radiomics}. Deep learning methods have achieved strong performance \cite{ronneberger2015u, chen2021transunet}, but require large-scale, high-quality annotations, which are costly and time-consuming, especially for 3D data \cite{jiao2024learning}. In contrast, unlabeled medical images are abundant, motivating research on semi-supervised medical image segmentation (SSMIS) to improve annotation efficiency.

\begin{figure}[t]
    \centering
    \includegraphics[width=0.48\textwidth]{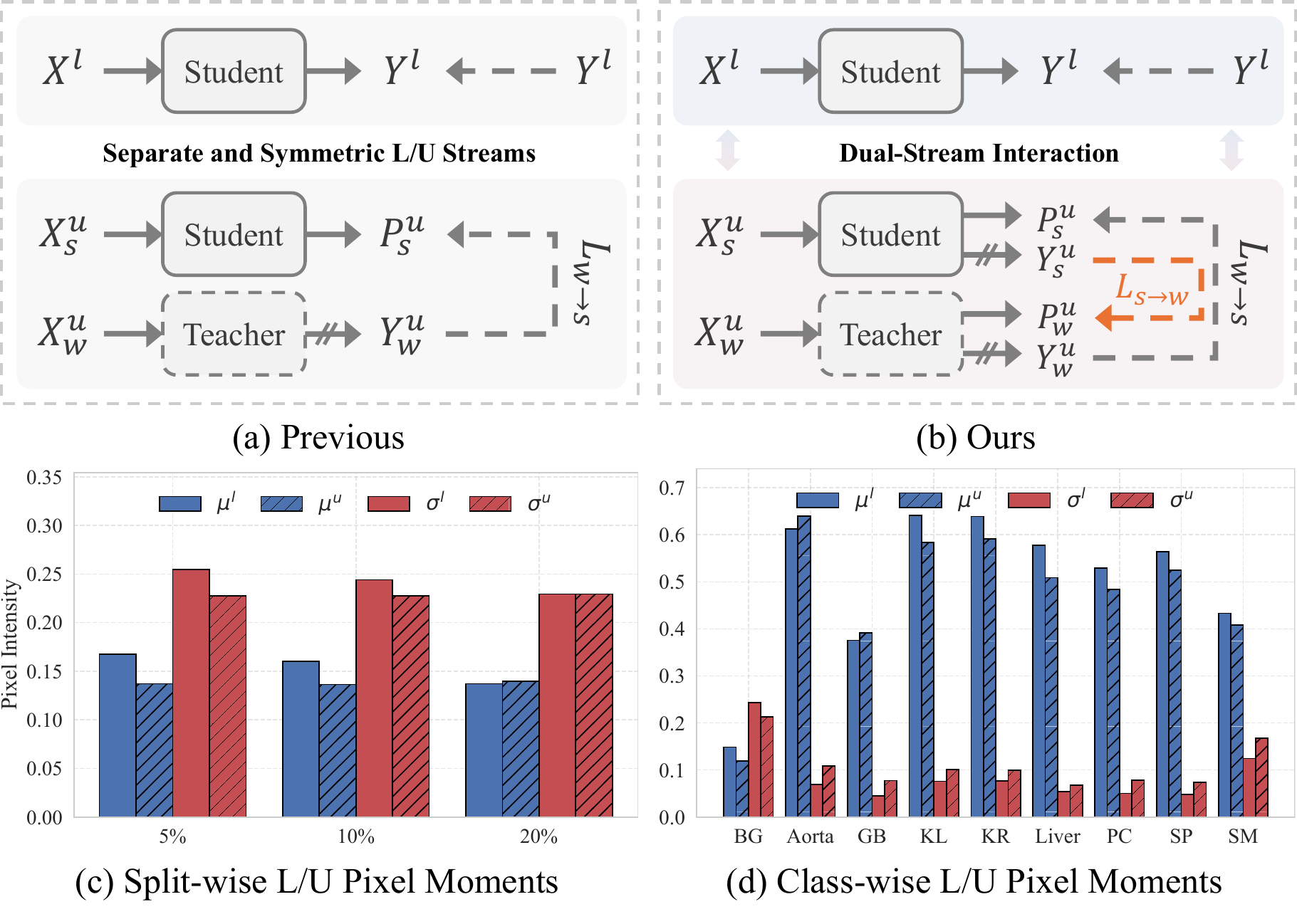}
    \caption{(a)–(b) show the previous (e.g., UA-MT \cite{yu2019uncertainty}) and proposed architectures, and (c)–(d) compare the first- and second-order moments of pixel intensities in the Synapse dataset \cite{landman2015miccai} under different data splits (c) and across categories (d).}
    \label{fig-1}
\end{figure}

Pseudo-labeling and weak-strong consistency regularization constitute prevalent strategies in SSMIS, wherein predictions from weakly perturbed unlabeled images serve as supervisory signals for their strongly perturbed counterparts, effectively guiding the decision boundary toward low-density regions \cite{yu2019uncertainty, jiang2024ph}. Conventionally, labeled and unlabeled data are processed through separate symmetric streams, as exemplified by AD-MT \cite{zhao2024alternate} and ABD \cite{chi2024adaptive}. However, the limited quantity of labeled data frequently fails to represent the underlying distribution comprehensively, inducing empirical distribution mismatch and confirmation bias \cite{wang2019semi, wang2024allspark}. To address this, cross-stream information interaction has been proposed, such as patch-wise augmentation in BCP \cite{bai2023bidirectional} and cross-attention feature translation in AllSpark \cite{wang2024allspark}, although these approaches remain constrained in resolving domain-specific intra-class distribution disparities \cite{qiu2024devil}. Within weak-strong consistency regularization, symmetric supervision, wherein predictions from both weak and strong augmentations provide mutual signals, is theoretically desirable \cite{chen2021exploring}. Nevertheless, the direct utilization of strong predictions as pseudo-labels is often eschewed due to the potential introduction of noise and consequent performance degradation \cite{wu2022exploring}. Recent SSMIS studies have incorporated contrastive learning to mitigate the limitations of pixel-wise consistency regularization; however, these approaches encounter challenges including dependence on labeled features \cite{zhang2023self}, susceptibility to noisy negative sampling \cite{jiang2024ph}, and the computational burden of point-to-point contrastive strategies \cite{liu2023multi}, collectively constraining their efficacy under conditions of extreme label scarcity \cite{li2024density}.

To address these limitations, we propose a \textbf{style-aware blending and prototype-based cross-contrast consistency learning framework} for semi-supervised medical image segmentation, illustrated in Fig.~\ref{fig-1}(b). The framework aims to establish an effective information interaction mechanism tailored to SSMIS. Empirical analyses (Fig.~\ref{fig-1}(c-d)) reveal that distributional discrepancies between labeled and unlabeled data are reflected in the first- and second-order statistical moments of pixel intensities, with disparities exacerbated under limited labeled data and varying across categories. Motivated by these observations, we introduce a \textbf{style-guided distribution blending} module to facilitate information exchange between labeled and unlabeled streams. Leveraging the enhanced generalization of this module alongside the robust capabilities of prototype-based contrastive learning \cite{zhang2023self, li2024density}, we further incorporate a \textbf{prototype-driven cross-contrast strategy} to reinforce weak-strong consistency regularization. Specifically, category-specific prototypes are computed from features of both weakly and strongly augmented views, followed by a cross-view point-to-prototype contrastive consistency mechanism that promotes reciprocal learning and refines representations. Extensive experiments demonstrate that the proposed framework achieves competitive performance, establishing new state-of-the-art results across multiple benchmarks.

\section{METHODS}\label{methods}

\begin{figure*}[ht]
    \centering
    \includegraphics[width=0.95\textwidth]{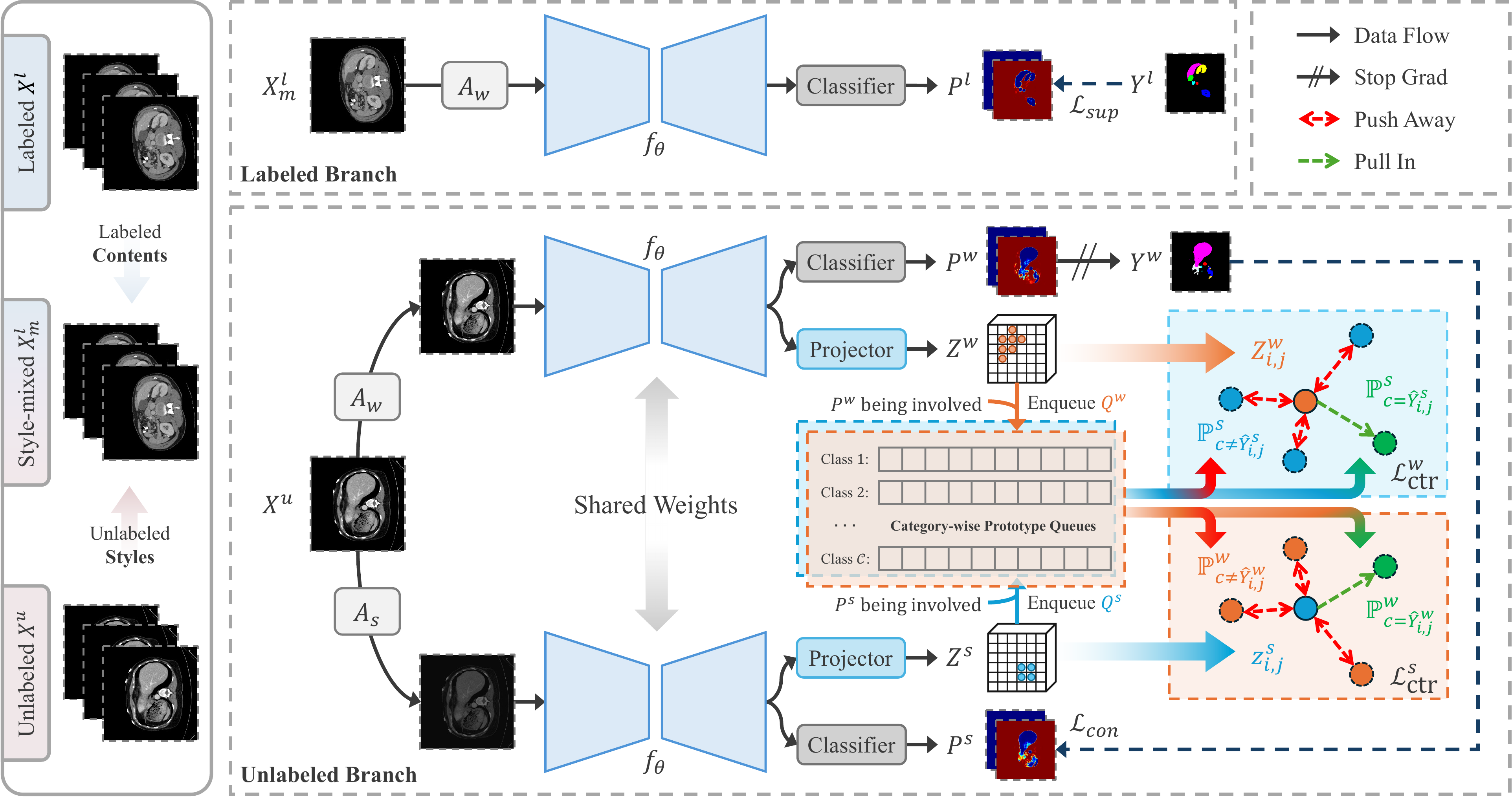}
    \caption{
        Illustration of our framework, comprising two main components: a style-guided distribution blending module (solid box) and a dual-branch architecture (dashed box) with labeled and unlabeled branches. The labeled branch is trained on style-blended labeled data, while the unlabeled branch enforces both pixel-wise weak-strong consistency and prototype-based cross-contrast consistency. Here, $X_m^l$ and $Y^l$ denote the style-blended labeled image and its corresponding label; $X^u$ and $Y^w$ represent the unlabeled image and its pseudo-label; $A_w$ and $A_s$ indicate weak and strong augmentations, respectively.
    }
    \label{fig-2}
\end{figure*}

\subsection{Problem Setting}\label{problem-setting}
In SSMIS, the dataset comprises a small labeled subset $\mathcal{D}^l=\{(X_i^l, Y_i^l)\}_{i=1}^{N}$ and a larger unlabeled subset $\mathcal{D}^u=\{X_i^u\}_{i=N+1}^{N+M}$, with $M \gg N$. Here, $X_i^l \in \mathbb{R}^{{H}\times{W}}$ is a labeled 2D image slice, ${Y_i^l}\in{\{0,1,\ldots,\mathcal{C}-1\}^{H \times W}}$ its segmentation mask, and $X_i^u$ an unlabeled slice. The subscript $i$ is omitted hereafter for simplicity.

\subsection{Overview}\label{overview}
Our overall framework follows the Mean Teacher \cite{tarvainen2017mean} paradigm, enhanced with weak-strong consistency regularization. As presented in Fig.~\ref{fig-2}, the architecture primarily comprises a style-guided distribution blending module and two parallel branches.

\textbf{Labeled Branch.} A labeled image $X^l$ first undergoes weak augmentations $A_w$, yielding a slightly perturbed image $A_w(X^l)$. This augmented image is then integrated with unlabeled images via the proposed style-guided distribution blending module to generate a mixed image $X_m^l$, as detailed in Section~\ref{style-guided-distribution-blending}. Subsequently, the resultant style-blended image is fed into the student model $f_\theta$ to produce the prediction $P^l$. Consistent with prior work \cite{bai2023bidirectional, chi2024adaptive}, we adopt the combination of cross-entropy and Dice loss as the supervised loss:
\begin{equation}
\mathcal{L}_{\text{sup}} = 
\frac{1}{2} \left( \mathcal{L}_{\text{ce}}(P^l, Y^l) + 
\mathcal{L}_{\text{dice}}(P^l, Y^l) \right).
\label{eq-1}
\end{equation}

\textbf{Unlabeled Branch.} To leverage unlabeled data, we employ a teacher model identical to the student model to generate pseudo-labels. An unlabeled image $X^u$ is first augmented with both weak $A_w$ and strong $A_s$ transformations, resulting in two views, $A_w(X^u)$ and $A_s(X^u)$. The teacher model processes the weakly augmented image to produce $P_w^u$, which serves as a pseudo-label to supervise the student model's prediction $P_s^u$ on the strongly augmented image. To fully exploit weak-strong consistency regularization, a consistency loss is combined with a prototype-based contrastive loss (Section~\ref{prototype-based-cross-contrast-consistency}) to define the unsupervised objective:
\begin{equation}
\mathcal{L}_{\text{unsup}} =
\alpha \, 
\mathcal{L}_{\text{dice}}\Big(P_s^u, \; \mathop{\arg\max}_{c \in \mathcal{C}} (P_{w}^u)_c \Big)
+ \beta \, \mathcal{L}_{\text{ctr}} \,,
\end{equation}
where $\alpha$ and $\beta$ control the contributions of the consistency and contrastive losses, respectively.

\textbf{Overall Objective.} The student model is optimized through gradient descent on the total loss:
\begin{equation}
\mathcal{L} = 
\mathcal{L}_{\text{sup}} + \lambda(t) \mathcal{L}_{\text{unsup}},
\end{equation}
where $\lambda(t)$ is a time-dependent weighting function, typically implemented as a Gaussian ramp-up function~\cite{tarvainen2017mean}.

\subsection{Style-guided Distribution Blending}\label{style-guided-distribution-blending}
To mitigate the confirmation bias induced by the dominant influence of the labeled stream, we begin by empirically analyzing the distributional discrepancies between labeled and unlabeled medical images. As illustrated in Fig.~\ref{fig-1}(c-d), these discrepancies are partially attributed to inconsistencies in the first- and second-order statistical moments of pixel intensities, which correspond to brightness and contrast in digital grayscale images \cite{haralick2007textural}. Such a pattern is evident in datasets such as Synapse \cite{landman2015miccai}, where variations in CT scanners and imaging protocols lead to distinct intensity statistics across anatomical regions. Motivated by this observation, a straightforward intervention is to explicitly blend the first- and second-order statistics derived from both domains to bridge this distributional disparity.

Inspired by \cite{huang2017arbitrary}, we propose a style-guided distribution blending module that injects diverse stylistic cues into labeled images while preserving their semantic content. For each sample in both batches, the corresponding style representations are computed as follows:
\begin{equation}
(\mu^{*}, \sigma^{*}) =
\left(
\mathbb{E}_{i,j}[X_{i,j}^{*}], \;
\sqrt{\mathbb{E}_{i,j}\!\left[(X_{i,j}^{*} - \mu^{*})^2\right] + \epsilon}
\right) \,,
\end{equation}
where the superscript $^*$ indicates either labeled or unlabeled samples, and $\epsilon$ is a small constant for numerical stability.

To generate training data covering both domains, a candidate style $(\mu^u, \sigma^u)$ is randomly sampled from the unlabeled style set and applied to the labeled content to produce a style-blended labeled batch:
\begin{equation}
X_m^l = \eta \, X^l + (1 - \eta) \left( \frac{X^l - \mu^l}{\sigma^l} \, \sigma^u + \mu^u \right) \,,
\end{equation}
where $\eta$ determines the relative contribution of labeled and unlabeled styles in the blended image.

Due to the limited style diversity of labeled data and to mitigate excessive information loss, we follow a strategy similar to that proposed in \cite{wang2024allspark}, where the style-mixed labeled batch is directly fed into the model for training.

\subsection{Prototype-based Cross-Contrast Consistency}\label{prototype-based-cross-contrast-consistency}
Existing consistency regularization frameworks \cite{chi2024adaptive, zhao2024alternate} often fail to enforce mutual consistency between predictions from different augmentations, as they neglect informative signals from strongly augmented outputs and direct mutual pseudo-labeling can introduce harmful noise. To rectify this, we introduce a prototype-based cross-contrast learning strategy to enforce mutual consistency while mitigating the impact of noisy predictions.

\begin{table*}[htbp]
\centering
\caption{Comparison with existing methods on the Synapse multi-organ dataset.}
\begin{tabularx}{\textwidth}{c|c|cc|*{8}{Y}}
\hline
Method & Labeled & DSC $\uparrow$ & ASD $\downarrow$ & Aorta & GB & KL & KR & Liver & PC & SP & SM \\
\hline
UNet\cite{ronneberger2015u} & 1(5\%) & 32.16 & 41.57 & 50.72 & 18.78 & 27.72 & 28.88 & 73.78 & 9.60 & 36.85 & 10.98 \\
UNet\cite{ronneberger2015u} & 2(10\%) & 37.75 & 43.54 & 58.06 & 23.07 & 39.33 & 27.94 & 81.00 & 10.48 & 40.71 & 21.39 \\
UNet\cite{ronneberger2015u} & 12(All) & 77.09 & 11.19 & 88.26 & 59.46 & 82.97 & 77.79 & 91.35 & 58.26 & 83.93 & 74.68 \\
\hline
UA-MT\cite{yu2019uncertainty} & \multirow{8}{*}{1(5\%)} & 34.36 & 45.59 & 64.42 & 21.88 & 26.46 & 29.91 & 73.31 & 11.75 & 34.22 & 12.89 \\
SS-Net\cite{wu2022exploring} &  & 36.95 & 31.20 & 65.44 & 24.51 & 39.16 & 19.78 & 86.40 & 2.04 & 51.19 & 7.11 \\
BCP\cite{bai2023bidirectional} &  & 43.09 & 62.03 & 65.31 & 12.44 & 48.18 & 46.25 & 82.79 & 8.51 & 62.34 & 18.88 \\
MCSC\cite{liu2023multi} & & 34.00 & -- & 50.90 & 13.00 & 17.60 & 54.60 & 64.30 & 5.50 & 43.10 & 23.50 \\
SCP-Net\cite{zhang2023self} &  & 36.38 & 40.00 & 58.54 & 21.06 & 34.92 & 35.84 & 76.74 & 11.43 & 39.48 & 13.04 \\
ABD\cite{chi2024adaptive} &  & 53.37 & \underline{18.89} & 74.87 & 2.73 & \underline{73.51} & \underline{70.69} & 89.56 & 17.10 & 63.77 & 34.69 \\
AD-MT\cite{zhao2024alternate} &  & 57.97 & 22.82 & 76.96 & 26.90 & \textbf{80.36} & \textbf{72.67} & 84.34 & 17.33 & 76.89 & 28.29 \\
\cline{3-12}
\textbf{Ours} (MT) & & \underline{60.00} & 19.73 & \underline{79.53} & \textbf{40.52} & 68.59 & 63.05 & \underline{89.80} & \underline{20.70} & \underline{78.38} & \underline{39.00} \\
\textbf{Ours} (BCP) & & \textbf{61.87} & \textbf{18.81} & \textbf{80.07} & \underline{38.78} & 68.30 & 63.12 & \textbf{91.60} & \textbf{24.54} & \textbf{80.65} & \textbf{47.93} \\
\hline
UA-MT\cite{yu2019uncertainty} & \multirow{8}{*}{2(10\%)} & 39.94 & 42.24 & 69.96 & 26.67 & 44.14 & 27.30 & 83.46 & 4.36 & 44.72 & 18.90 \\
SS-Net\cite{wu2022exploring} & & 41.46 & 24.50 & 77.94 & 31.54 & 54.01 & 12.02 & 86.25 & 3.36 & 56.98 & 9.55 \\
BCP\cite{bai2023bidirectional} & & 51.77 & 43.11 & 45.08 & 31.34 & 62.03 & 58.05 & \textbf{91.07} & 16.34 & 78.82 & 31.44 \\
MCSC\cite{liu2023multi} &  & 61.10 & -- & 73.90 & 26.40 & 69.90 & 72.70 & 90.00 & 33.20 & \underline{79.40} & 43.00 \\
SCP-Net\cite{zhang2023self} & & 45.07 & 25.62 & 63.44 & 25.81 & 57.95 & 33.07 & 89.64 & 14.70 & 52.99 & 22.99 \\
ABD\cite{chi2024adaptive} & & 59.61 & \textbf{9.47} & 78.64 & 4.93 & 73.27 & 68.91 & 90.23 & \underline{36.97} & 76.44 & 47.48 \\
AD-MT\cite{zhao2024alternate} & & 60.47 & 20.63 & 69.55 & 28.72 & \underline{76.44} & \underline{74.56} & 89.18 & 30.53 & \textbf{80.86} & 33.89 \\
\cline{3-12}
\textbf{Ours} (MT) & & \underline{65.81} & \underline{15.88} & \underline{80.77} & \textbf{45.42} & 76.39 & \textbf{76.45} & 90.10 & 30.14 & 75.84 & \underline{51.36} \\
\textbf{Ours} (BCP) & & \textbf{66.59} & 23.52 & \textbf{81.79} & \underline{40.09} & \textbf{76.80} & 68.64 & \underline{90.43} & \textbf{38.80} & 75.79 & \textbf{60.39} \\
\hline
\end{tabularx}
\label{tab-1}
\end{table*}

\subsubsection{\textbf{Confidence-guided Prototype Estimation}}
To address the potential noise in pixel-level predictions, we propose a confidence-guided, category-wise feature aggregation strategy for prototype estimation. This design helps mitigate the incorrect sampling of positive and negative examples. Given an unlabeled image $X^u$, we first obtain its weakly and strongly augmented views, denoted as $A_w(X^u)$ and $A_s(X^u)$, respectively. The corresponding feature representations and class probability maps are computed as:
\begin{equation}
Z^{v} = q(f_\theta(A_{v}(X^u))) \in \mathbb{R}^{D \times H \times W} \,,
\label{eq-10}
\end{equation} 
\begin{equation}
P^v = \mathrm{softmax}(h(f_\theta(A_v(X^u)))) \in \mathbb{R}^{\mathcal{C} \times H \times W} \,,
\end{equation}
where $v \in \{w, s\}$ denotes weak or strong views. $f_\theta$ denotes the shared backbone, $q(\cdot)$ is a non-linear projection head composed of three stacked Conv-ReLU-BN layers, and $h(\cdot)$ represents the segmentation head. $D$ and $\mathcal{C}$ represent the dimensions of the feature and class spaces, respectively.

The category-wise prototypes from weak and strong views, denoted as $\tilde{\mathbb{P}}_c^w$ and $\tilde{\mathbb{P}}_c^s$, are computed via confidence-weighted feature aggregation as follows:
\begin{equation}
\tilde{\mathbb{P}}_c^v =
\frac{\mathbb{E}_{i,j} \big[ 
      \underbrace{P_{c,i,j}^v}_{\text{pixel weight}} 
      \odot 
      \underbrace{Z_{:,i,j}^v}_{\text{feature vector}} 
    \big]}
{\mathbb{E}_{i,j} \big[ P_{c,i,j}^v \big]} \,, \quad c = 1, \dots, \mathcal{C}.
\end{equation}

Ideally, the prototype of each category should capture the overall feature distribution of the dataset in the embedding space. However, due to factors such as the absence of certain categories in a batch, limited batch size, and uneven learning progress across classes \cite{he2020momentum}, prototypes estimated solely from individual batches tend to be biased. To address this issue, we adopt a queue-based category-wise memory bank $[\tilde{\mathbb{P}}_{c_1}^{v_1}, \tilde{\mathbb{P}}_{c_2}^{v_2}, \dots, \tilde{\mathbb{P}}_{c_K}^{v_K}]$ of size $K$ that caches prototypes generated during sequential training. The memory bank is updated in a First-In-First-Out (FIFO) manner, and the final prototype representation for each category is computed as:
\begin{equation}
\mathbb{P}_c^v =
\frac{\sum_{k=1}^{K} 
      \underbrace{\mathbb{E}_{i,j}[P_{c_k,i,j}^{v_k}]}_{\text{batch weight}} 
      \times
      \underbrace{\tilde{\mathbb{P}}_{c_k}^{v_k}}_{\text{prototype}}}
{\sum_{k=1}^{K} \mathbb{E}_{i,j}[P_{c_k,i,j}^{v_k}]} \,, \quad c = 1, \dots, \mathcal{C}.
\end{equation}

\subsubsection{\textbf{Cross-Contrast Consistency Loss}}
As shown in Fig.~\ref{fig-2}, the rationale behind applying both weak and strong augmentations simultaneously to the same unlabeled image is to leverage the inherent consistency expected between their predicted semantics. This mutual consistency allows predictions from one view to serve as supervisory signals for the other, encouraging intra-class features to cluster together while pushing inter-class features apart in the embedding space. To this end, we employ a cross-view contrastive learning strategy to enforce mutual consistency: pixel-level features from the weakly augmented branch are drawn toward class prototypes from the strongly augmented branch and repelled from dissimilar ones, and vice versa. The contrastive loss is defined accordingly:
\begin{equation}
\mathcal{L}_{\mathrm{ctr}} =
\frac{1}{2} \sum_{v \in \{w,s\}}
- \log \frac{
\exp\big(\mathrm{sim}(Z_{i,j}^{v}, \mathbb{P}_{\hat{Y}_{i,j}^{v}}^{\bar{v}})/\tau\big)
}{
\sum_{c=1}^{\mathcal{C}} \exp\big(\mathrm{sim}(Z_{i,j}^{v}, \mathbb{P}_c^{\bar{v}})/\tau\big)
} \,.
\end{equation}
Here, $\tau$ is a temperature hyperparameter. $\hat{Y}_{i,j}^v$ denotes the prediction produced by the EMA-updated model $f_{\bar{\theta}}$ of the student network, whose parameters are updated as $\bar{\theta} \leftarrow \gamma \cdot \bar{\theta} + (1 - \gamma) \cdot \theta$, where $\gamma = 0.99$.

\begin{table}[t]
\centering
\caption{
Comparison with existing methods on the ACDC dataset.
}
\begin{tabularx}{\columnwidth}{c|c|cc|*{3}{Y}}
\hline
Method & Labeled & DSC $\uparrow$ & ASD $\downarrow$ & Myo & LV & RV \\
\hline
UNet\cite{ronneberger2015u} & 3(5\%) & 48.90 & 13.65 & 37.13 & 51.30 & 58.27 \\
UNet\cite{ronneberger2015u} & 7(10\%) & 80.75 & 2.75 & 74.84 & 80.74 & 86.67 \\
UNet\cite{ronneberger2015u} & 70(All) & 91.64 & 0.45 & 90.59 & 89.39 & 94.95 \\
\hline
UA-MT\cite{yu2019uncertainty} & \multirow{8}{*}{3(5\%)} & 56.58 & 8.04 & 41.40 & 59.10 & 69.24 \\
SS-Net\cite{wu2022exploring} & & 65.82 & 2.28 & 57.55 & 65.66 & 74.26 \\
BCP\cite{bai2023bidirectional} & & 87.59 & \underline{0.68} & 85.71 & \underline{85.97} & 91.09 \\
MCSC\cite{liu2023multi} & & 73.60 & -- & 70.00 & 79.20 & 71.70 \\
SCP-Net\cite{zhang2023self} & & 70.93 & 6.55 & 61.89 & 70.78 & 80.11 \\
ABD\cite{chi2024adaptive} & & 85.15 & 2.81 & 84.46 & 82.71 & 88.29 \\
AD-MT\cite{zhao2024alternate} & & \underline{88.22} & 0.94 & \underline{86.68} & \textbf{86.13} & \underline{91.85} \\
\cline{3-7}
\textbf{Ours} (MT) & & 87.31 & 1.10 & 85.60 & 84.90 & 91.42 \\
\textbf{Ours} (BCP) & & \textbf{88.60} & \textbf{0.61} & \textbf{87.36} & 85.51 & \textbf{92.93} \\
\hline
UA-MT\cite{yu2019uncertainty} & \multirow{8}{*}{7(10\%)} & 80.60 & 2.91 & 77.87 & 78.79 & 85.13 \\
SS-Net\cite{wu2022exploring} & & 86.78 & 1.40 & 85.36 & 84.34 & 90.64 \\
BCP\cite{bai2023bidirectional} & & 88.84 & 1.17 & 87.68 & 86.54 & 92.30 \\
MCSC\cite{liu2023multi} & & \underline{89.40} & -- & 87.60 & \textbf{93.60} & 87.10 \\
SCP-Net\cite{zhang2023self} & & 88.17 & 1.67 & 87.74 & 85.39 & 91.39 \\
ABD\cite{chi2024adaptive} & & 87.62 & 3.08 & 88.19 & 84.50 & 90.17 \\
AD-MT\cite{zhao2024alternate} & & 89.07 & \underline{0.82} & 88.12 & \underline{87.11} & 91.96 \\
\cline{3-7}
\textbf{Ours} (MT)  & & 89.27 & \textbf{0.63} & \underline{88.35} & 86.63 & \underline{92.82} \\
\textbf{Ours} (BCP) & & \textbf{89.80} & 0.95 & \textbf{90.36} & 86.10 & \textbf{92.94} \\
\hline
\end{tabularx}
\label{tab-2}
\end{table}

\section{EXPERIMENTS AND RESULTS}
\subsection{Datasets and Implementation Details}
We evaluate the proposed method on two public benchmarks. The \textbf{ACDC dataset} \cite{bernard2018deep} comprises 100 cardiac MRI scans annotated with three classes: left ventricle (LV), myocardium (Myo), and right ventricle (RV). Following \cite{luo2022semi}, 70 cases (1312 slices) are used for training, 10 for validation, and 20 for testing under a 7:1:2 split. For the semi-supervised setting, 5\% (3 cases) and 10\% (7 cases) of the training data are labeled. The \textbf{Synapse dataset} \cite{landman2015miccai} includes 30 abdominal CT volumes annotated with eight organs: aorta, gallbladder (GB), spleen (SP), left kidney (KL), right kidney (KR), liver, pancreas (PC), and stomach (SM). Following \cite{chen2021transunet}, 18 cases (2211 slices) are used for training and 12 for testing, with 5\% (1 case) and 10\% (2 cases) of the training data labeled. The model is trained for $30$k iterations using stochastic gradient descent (SGD) with a learning rate of $0.01$, weight decay of $10^{-4}$, and batch size of $24$. All images are resized to $256 \times 256$. Weak augmentations include random rotation and flipping, while strong augmentations employ ColorJitter and Gaussian blur. Hyperparameters are set as: weighting coefficients $\alpha=0.1$, $\beta=1.0$, style-mixing coefficient $\eta \sim \mathcal{U}(0,1)$, temperature $\tau=1.0$, and memory bank size $K=128$.

\begin{table}[t]
\caption{
Ablation study on the effect of different components. \texttt{Base}: baseline. \texttt{SDB}: Style-guided Distribution Blending. $\mathcal{L}_{\mathrm{ctr}}$: Prototype-based Cross-contrast Loss. \texttt{MB}: Memory Bank.
}
\begin{tabularx}{\columnwidth}{Y|YYYY|YY}
\hline
\multirow{2}{*}{Dataset} & \multicolumn{4}{c|}{Components} & \multirow{2}{*}{DSC $\uparrow$} & \multirow{2}{*}{ASD $\downarrow$} \\
\cline{2-5}
& \raisebox{-0.1\height}[0pt]{Base} & \raisebox{-0.1\height}[0pt]{SDB} & \raisebox{-0.1\height}[0pt]{$\mathcal{L}_{\mathrm{ctr}}$} & \raisebox{-0.1\height}[0pt]{MB} & & \\
\hline
\multirow{5}{*}{\shortstack{Synapse \\ (10\%)}} & \checkmark & & & & 55.78 & 25.91 \\
 & \checkmark & \checkmark & & & 62.11 & 19.28 \\
 & \checkmark & & \checkmark & & 61.28 & 23.03 \\
 & \checkmark & \checkmark & \checkmark & & 64.23 & 17.36 \\
 & \checkmark & \checkmark & \checkmark & \checkmark & \textbf{65.81} & \textbf{15.88} \\
\hline
\multirow{5}{*}{\shortstack{ACDC \\ (10\%)}} & \checkmark & & & & 85.84 & 1.58 \\
 & \checkmark & \checkmark & & & 87.91 & 1.04 \\
 & \checkmark & & \checkmark & & 87.77 & 0.99 \\
 & \checkmark & \checkmark & \checkmark & & 88.38 & 0.97 \\
 & \checkmark & \checkmark & \checkmark & \checkmark & \textbf{89.27} & \textbf{0.63} \\
 \hline
\end{tabularx}
\label{tab-3}
\end{table}

\subsection{Comparison with Existing Methods}
\textbf{Synapse dataset.} Table~\ref{tab-1} shows that our method outperforms recent state-of-the-art approaches on the Synapse dataset under both 5\% and 10\% settings, achieving higher mean DSC and competitive ASD. The BCP-extended framework yields notable mean DSC gains of 18.78\% and 14.82\% over BCP, confirming its effectiveness in exploiting intermediate domains and informative supervision. It also achieves substantial improvements in challenging organs (e.g., +20.12\% and +19.92\% on the stomach), and under the 5\% setting surpasses distribution-mixing methods such as BCP across multiple categories (e.g., +7.01\% on the liver), demonstrating the advantage of the proposed style-guided distribution blending. Moreover, our method consistently exceeds weak-strong consistency methods (ABD, AD-MT) and contrastive learning approaches (MCSC, SCP-Net), with mean DSC improvements of 4.71\% and 20.74\% in the 10\% labeled setting.

\begin{table}[!tbp]
\caption{
Ablation study of each cross-contrast consistency loss. $\mathcal{L}_{\mathrm{pixel}}^s$: strong-to-weak Dice loss. $\mathcal{L}_{\mathrm{ctr}}^w$ / $\mathcal{L}_{\mathrm{ctr}}^s$: weak-to-strong / strong-to-weak contrastive loss.
}
\begin{tabularx}{\columnwidth}{YYYY|YY}
\hline
w/ SDB & $\mathcal{L}_{\mathrm{pixel}}^s$ & $\mathcal{L}_{\mathrm{ctr}}^w$ & $\mathcal{L}_{\mathrm{ctr}}^s$ & DSC $\uparrow$ & ASD $\downarrow$ \\
\hline
\checkmark & & & & 62.11 & 19.28 \\
\checkmark & \checkmark & & & 61.29 & 24.56 \\
\checkmark & & \checkmark & & 62.35 & 16.34 \\
\checkmark & & & \checkmark & 62.14 & 21.32 \\
\checkmark & & \checkmark & \checkmark & \textbf{65.81} & \textbf{15.88} \\
\hline
\end{tabularx}
\label{tab-5}
\end{table}

\textbf{ACDC dataset.} Table~\ref{tab-2} shows that our method consistently achieves the highest average performance on the ACDC dataset with both 5\% and 10\% labeled splits, improving BCP by 1.01\% when integrated and outperforming MCSC by 13.71\% under 5\%, highlighting its scalability and effectiveness in bridging labeled-unlabeled gaps and addressing distribution shifts. Even at 10\%, it surpasses contrastive methods like SCP-Net by 1.1\%, demonstrating superior utilization of strongly augmented predictions.

\begin{figure}[t]
\centering
\includegraphics[width=0.48\textwidth]{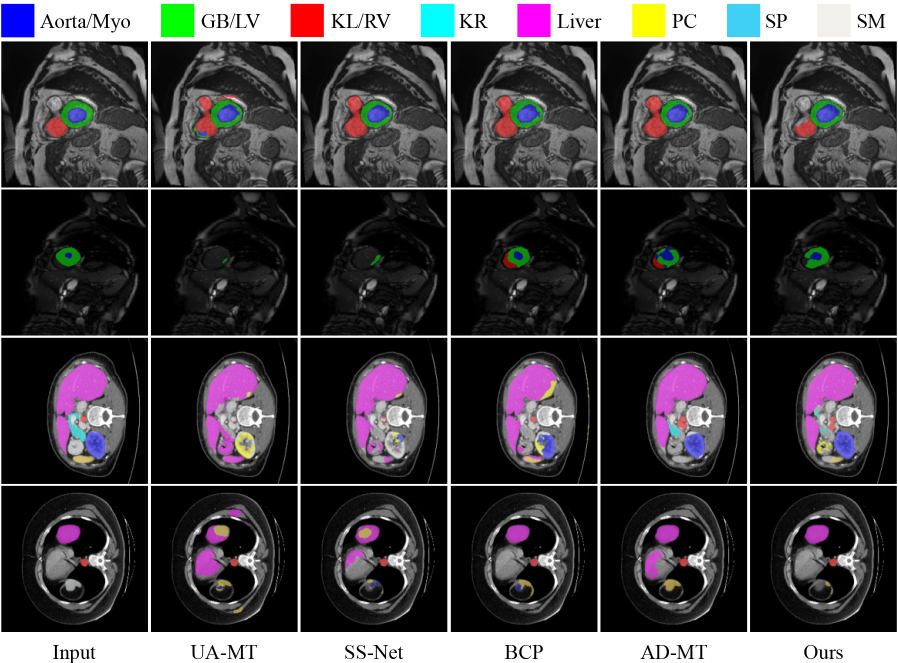}
\caption{
Visual segmentation results of different methods on ACDC (top two rows) and Synapse (bottom two rows). Our method achieves a better balance between under-segmentation and over-segmentation.
}
\label{fig-3}
\end{figure}

\textbf{Qualitative Analysis.} To visually substantiate the superiority of our method, Fig.~\ref{fig-3} presents representative segmentation results on the ACDC and Synapse datasets. On ACDC, our approach achieves a better balance between under- and over-segmentation, particularly for the right ventricle, while producing more precise boundaries and higher recall across other classes. On Synapse, it demonstrates superior delineation of organs such as the aorta and liver and yields more reliable segmentation for challenging structures like the pancreas.

\subsection{Ablation Studies}
A stepwise ablation study evaluates the contribution of each component (Tables~\ref{tab-3} and~\ref{tab-5}). The baseline achieves DSCs of 55.78\% (Synapse) and 85.84\% (ACDC). Incorporating the style-guided distribution blending (SDB) improves DSC, and adding the cross-contrast loss ($\mathcal{L}_{\mathrm{ctr}}$) further enhances performance, with the full framework attaining DSCs of 65.81\% (Synapse) and 89.27\% (ACDC) and ASDs of 15.88mm and 0.63mm, corresponding to absolute DSC gains of 10.03\% and 3.43\% over the baseline. Analysis of the cross-contrast consistency loss components shows that a pixel-level Dice loss from strong to weak views ($\mathcal{L}_{\mathrm{pixel}}^s$) degrades performance (DSC -0.82\%, ASD +5.28mm), indicating that direct supervision from noisy predictions is detrimental. In contrast, $\mathcal{L}_{\mathrm{ctr}}^w$ and $\mathcal{L}_{\mathrm{ctr}}^s$ individually improve segmentation, and their combination achieves the best results, confirming the complementary benefit of bidirectional contrastive consistency.

\section{CONCLUSION}
In this paper, we propose a style-aware blending and prototype-based cross-contrast consistency learning framework to improve semi-supervised medical image segmentation. The central idea is to further leverage weak-strong consistency regularization under specific perturbations, thereby pushing the upper bound of consistency learning. To tackle the prevalent challenges of distribution mismatch and insufficient exploitation of supervision signals in semi-supervised segmentation, we introduce a style-guided distribution blending module for fine-grained alignment and a prototype-based cross-contrast module to enable bidirectional supervision. Through the collaborative design of these two modules, our framework not only achieves state-of-the-art performance on two benchmark datasets but also enhances existing semi-supervised segmentation methods from the perspectives of style adaptation and supervision efficacy, as validated by extensive ablation studies.

\section*{Acknowledgment}
This work was supported by the National Natural Science Foundation of China (62462068).

\small
\bibliography{references.bib}
\bibliographystyle{IEEEtran}
\end{document}